\documentclass[journal]{IEEEtran} 

\IEEEoverridecommandlockouts                   
\usepackage{graphicx}
\usepackage{caption}
\usepackage{subcaption}
\usepackage{url}
\usepackage{xcolor}
\usepackage{hyperref}
\usepackage{amssymb}
\usepackage{booktabs}
\usepackage{tabularx}
\usepackage{tikz}
\usepackage{microtype} 
\usepackage{pifont}
\usepackage{xcolor}
\usepackage{booktabs} 
\usepackage{siunitx} 
\usepackage{multirow}

\usepackage[printonlyused,nolist,withpage,nohyperlinks]{acronym} 
\begin{acronym}

\acro{RL}{reinforcement learning}
\acro{DRL}{deep reinforcement learning}
\acro{SOTA}{state-of-the-art}
\acro{SB3}{stable-baselines3}
\acro{Sim2Real}{simulation to real transfer}
\acro{MDP}{Markov Decision Process}
\acro{API}{Application Programming Interface}

\acro{C-Space}{configuration space}
\acro{RPP}{randomized path planner}
\acro{PRM}{Probabilistic Roadmap}
\acro{Lazy PRM}{Lazy PRM}
\acro{RRT}{Rapidly-exploring Random Tree}
\acro{RRT-Connect}{RRT-connect}
\acro{RRT*}{asymptotically optimal RRT}
\acro{T-RRT}{transition-based RRT}
\acro{CL-RRT}{closed-Loop RRT}
\acro{RRG}{rapidly-exploring random graph}
\acro{PMP}{partial motion planner}
\acro{CPP}{coverage path planning}
\acro{AFP}{artificial potential field}
\acro{CHOMP}{Covariant Hamiltonian Optimization for Motion Planning}

\acro{AD*}{anytime dynamic A*}

\acro{2D}{2-dimensional}
\acro{3D}{3-dimensional}
\acro{EST}{expansive-spaces tree}

\acro{FM}{fast marching}

\acro{OMPL}{Open Motion Planning Library}
\acro{FCL}{flexible collision library}
\acro{WP}{waypoint}

\acro{B-spline}{basis spline}

\acro{HLP}{high-level planner}
\acro{LLP}{low-level planner}

\acro{CC}{chance constraints}
\acro{DFS}{depth-first search}

\acro{DOF}{degrees of freedom}
\acro{ROS}{Robot Operating System}
\acro{SLAM}{simultaneous localization and mapping}
\acro{UAV}{unmanned aerial vehicle}
\acro{UGV}{unmanned ground vehicle}
\acro{GNC}{guidance, navigation and control}
\acro{GA}{genetic algorithm}
\acro{GPS}{global positioning system}
\acro{INS}{inertial navigation sensors}
\acro{DR}{dead-reckoning}
\acro{SLAM}{Simultaneous Localization and Mapping}
\acro{URDF}{Unified Robot Description Format}

\acro{UUV}{unmanned underwater vehicle}
\acro{UV}{underwater vehicle}
\acro{ROV}{remotely operated vehicle}
\acrodefplural{ROV}[ROVs]{remotely operated vehicles}
\acro{AUV}{autonomous underwater vehicle}
\acrodefplural{AUV}[AUVs]{autonomous underwater vehicle}

\acro{USBL}{ultra-short baseline}
\acro{DVL}{doppler velocity log}
\acro{IMU}{inertial measurement unit}

\acro{LOS}{line-of-sight}
\acro{NED}{north-east-depth}

\acro{CIRS}{Underwater Vision and Robotics Research Center}
\acro{ViCOROB}{Computer Vision and Robotics Institute}
\acro{COLA2}{component oriented layer-based architecture for autonomy}
\acro{UWSim}{underwater simulator}
\acro{MSV}{MORPH supra-vehicle}

\acro{Ifremer}{French Research Institute for Exploitation of the Sea}
\acro{ACE}{automated control engine}
\acro{ISE}{International Submarine Engineering}
\acro{SDNC}{splats distance normalized cut}

\acro{MPC}{model predictive control}

\acro{FoV}{field of view}

\acro{GUI}{Graphical User Interface}

\end{acronym}

\begin{acronym}
    \acro{auv}[AUV]{Autonomous Underwater Vehicle}
    \acro{rov}[ROV]{Remotely Operated Vehicle}
    \acro{ds}[DS]{Docking Station}
    \acro{pid}[PID]{Proportional-Integral-Derivative Controller}
    \acro{ned}[NED]{North, East, Down}
    \acro{urdf}[URDF]{Unified Robotics Description Format}
    \acro{cad}[CAD]{Computer-aided Design}
    \acro{vicorob}[ViCOROB]{Computer Vision and Robotics}
    \acro{cirs}[CIRS]{ Centro de Investigación en Robótica Submarina}
    \acro{udg}[UDG]{Universitat de Griona}
\end{acronym}

\title{Advancing Shared and Multi-Agent Autonomy in Underwater Missions: Integrating Knowledge Graphs and Retrieval-Augmented Generation}

\def\BibTeX{{\rm B\kern-.05em{\sc i\kern-.025em b}\kern-.08em
    T\kern-.1667em\lower.7ex\hbox{E}\kern-.125emX}}

\makeatletter
\newcommand{\linebreakand}{%
  \end{@IEEEauthorhalign}
  \hfill\mbox{}\par
  \mbox{}\hfill\begin{@IEEEauthorhalign}
}

\author{Michele~Grimaldi$^*$, %
        Carlo~Cernicchiaro$^*$, 
        Sebastian~Realpe Rua$^*$, 
        Alaaeddine~El-Masri-El-Chaarani, Markus Buchholz,
        Loizos~Michael, 
        Pere~Ridao~Rodriguez, 
        Ignacio~Carlucho, %
        and~Yvan~R.~Petillot %
\thanks{\textsuperscript{*}These authors contributed equally.}
\thanks{M. Grimaldi, M. Buchholz, I. Carlucho, and Y.  R. Petillot are with the School of Engineering and Physical Sciences, Heriot-Watt University, Edinburgh, UK. e-mail: mg2084@hw.ac.uk.} %
\thanks{C. Cernicchiaro and L. Michael are with the Open University of Cyprus, Nicosia, Cyprus. }
\thanks{S. Realpe Rua, A. El-Masri-El-Chaarani, and P. Ridao Rodriguez are with the University of Girona, Girona, Spain.}}

\begin{document}

\maketitle

\begin{abstract}
Robotic platforms have become essential for marine operations by providing regular and continuous access to offshore assets, such as underwater infrastructure inspection, environmental monitoring, and resource exploration. However, the complex and dynamic nature of underwater environments—characterized by limited visibility, unpredictable currents, and communication constraints—presents significant challenges that demand advanced autonomy while ensuring operator trust and oversight. Central to addressing these challenges are knowledge representation and reasoning techniques, particularly knowledge graphs and retrieval-augmented generation (RAG) systems, that enable robots to efficiently structure, retrieve, and interpret complex environmental data. These capabilities empower robotic agents to reason, adapt, and respond effectively to changing conditions. 
The primary goal of this work is to demonstrate both multi-agent autonomy and shared autonomy, where multiple robotic agents operate independently while remaining connected to a human supervisor. We show how a RAG-powered large language model, augmented with knowledge graph data and domain taxonomy, enables autonomous multi-agent decision-making and facilitates seamless human-robot interaction, resulting in 100\% mission validation and behavior completeness. Finally, ablation studies reveal that without structured knowledge from the graph and/or taxonomy, the LLM is prone to hallucinations, which can compromise decision quality.
\end{abstract}

\begin{figure*}[t]
  \centering
  \includegraphics[width=0.95\linewidth]{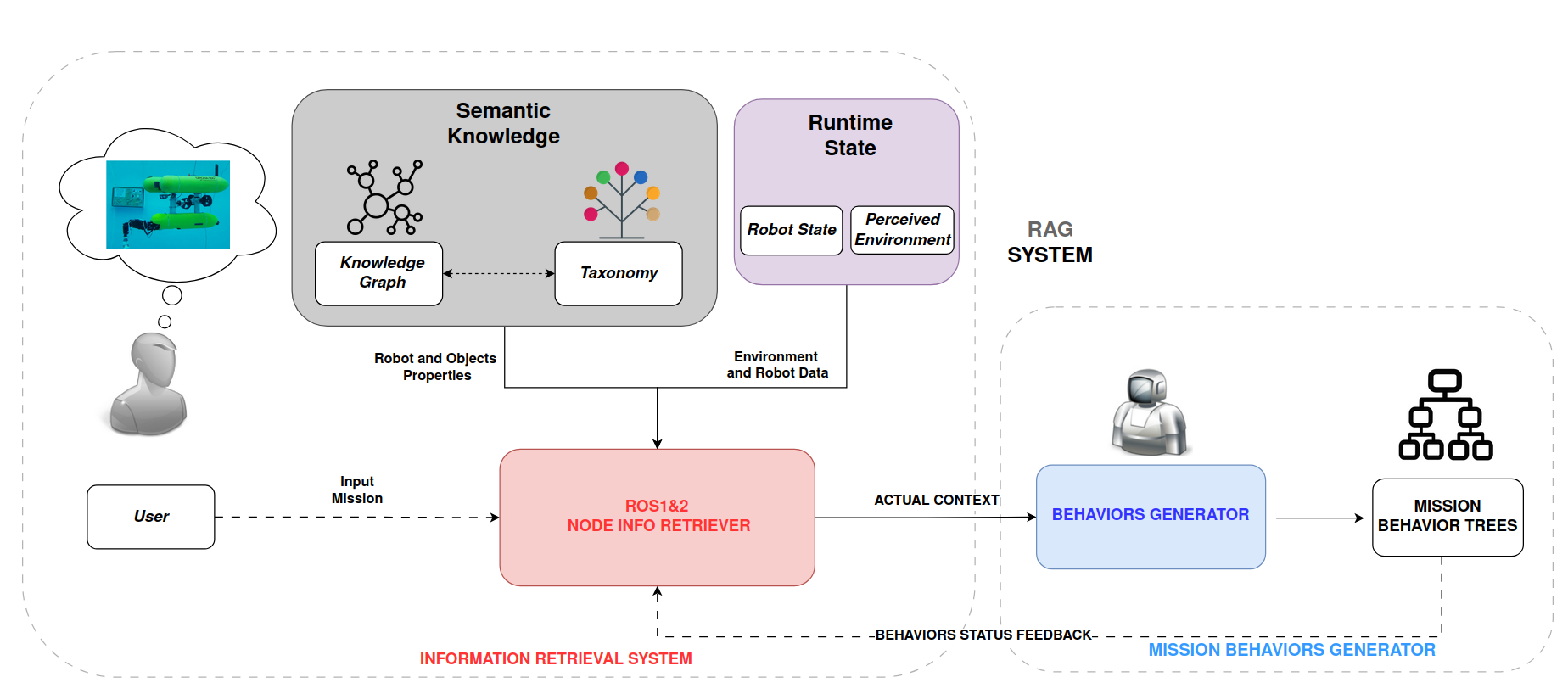}
\caption{Overview of the RAG system, which integrates an Information Retrieval System and a Mission Behaviors Generator. The user-defined mission is processed by the ROS Node Info Retriever, which queries semantic knowledge (Knowledge Graph and Taxonomy) and runtime state to obtain relevant robot and environment data. This information guides the Behaviors Generator in producing adaptive Mission Behavior Trees, enabling execution and real-time feedback-driven updates.}
  \label{fig:system_architecture}
\end{figure*}

\section{Introduction}

Marine operations increasingly rely on robotic platforms to perform tasks in challenging underwater environments. These platforms, such as \acp{AUV}~\cite{ROVsToAUVs, 6093749}, provide critical capabilities for industries like offshore wind, oil and gas, and marine research. The need for a regular and long-lasting presence in the underwater realm has driven the demand for continuous sensing and monitoring systems, which can operate in remote and difficult-to-reach locations for extended periods of time~\cite{ROVsToAUVs}. AUVs offer greater manoeuvrability, access to confined spaces, and the ability to operate at greater depths, which makes them particularly well-suited for applications such as deep-sea mapping, environmental monitoring, and resource exploration \cite{9463768}.

Much of the research in marine robotics has focused on advancing autonomy for AUVs to minimize reliance on continuous human supervision. State-of-the-art autonomy frameworks combine knowledge representation and reasoning, which let vehicles build environmental models, maintain situational awareness, and reason about goals and constraints \cite{maurelli2014cognitive, insaurralde2015capability}, with robust planning mechanisms that dynamically adapt actions to the unpredictable underwater domain. To extend mission endurance and data throughput, this algorithmic core is paired with hardware infrastructure such as autonomous docking stations, enabling AUVs to recharge and offload data between sorties \cite{PALOMERAS2018106,9440960}.

Despite these advances, achieving fully independent operation in complex marine environments remains an open challenge \cite{atyabi2018current}. Consequently, shared autonomy—in which decision authority is flexibly allocated between the vehicle and a remote operator—continues to play a crucial role in ensuring both safety and mission success. %
Additionally, the underwater domain presents unique challenges that significantly complicate the deployment and management of these robotic systems in a fully autonomous way. One of the primary obstacles is the limitation of underwater communication. Acoustic modems, the most common method for underwater communication, are constrained by a trade-off between range and bandwidth, where higher bandwidth results in shorter range and longer range requires lower bandwidth. Optical modems \cite{10244690}, while offering higher data rates, are constrained by their limited range and susceptibility to environmental conditions such as turbidity and light absorption \cite{grimaldi2023investigation}. 
On the other hand, by utilizing a shared autonomy approach, human operators can provide critical insights, context, and situational awareness, particularly in complex or ambiguous situations where robotic systems may struggle to adapt. This is particularly relevant in subsea operations, where traditional teleoperation using remotely operated vehicles (ROVs) often suffers from limited situational awareness and high operator workload. Recent efforts to address these limitations include the development of underwater Digital Twin (DT) systems, which integrate virtual reality, mapping, localization, and simulation to improve teleoperation and reduce cognitive demands on human operators~\cite{favor}.However, the unpredictable nature of underwater communication necessitates a degree of embedded autonomy as well. Autonomous decision-making, planning, and coordination become essential when communication with human operators is intermittent or unavailable. Therefore, a balanced approach that integrates both shared and embedded autonomy is crucial for ensuring effective performance in complex underwater missions.

In this article, we explore how knowledge representation, reasoning, and planning tools can facilitate the seamless interoperability of autonomous agents. The primary contribution of this work is the development of a framework designed to support mission planning in both shared and full autonomy for tasks such as surveying, inspection, and intervention, all while addressing the unique constraints presented by underwater environments. To improve decision-making and ensure precise information retrieval, we integrate a Large Language Model (LLM) with a Retrieval Augmented Generation (RAG) \cite{rag}, and a Knowledge Graph (KG), allowing exact match retrieval based on structured relationships. This approach ensures deterministic and interpretable query results, as the knowledge graph and the taxonomy encode domain-specific knowledge, capturing strict dependencies between concepts. A schema summarizing the interaction among all components is shown in Figure \ref{fig:system_architecture}.
We present a use case involving two AUVs, each with distinct roles. The AUVs are tasked with identifying and interacting with various objects to demonstrate the framework’s capabilities in a surveying and intervention mission, leveraging the robot’s knowledge in realistic environmental conditions.
Our results demonstrate that our method enables robots to perform complex missions in realistic simulated marine environments, allowing not only the coordination of multiple robotic platforms but also complex human-robot interaction.   
Furthermore, our results show that our framework achieves 100\% accuracy in mission execution by leveraging the knowledge graph and taxonomy. This work represents a significant milestone in marine robotics, demonstrating for the first time how LLMs can enable shared autonomy in underwater environments and paving the way for real-world deployment of autonomous underwater systems.

\section{Related Work}

Autonomous underwater missions require robust knowledge representation and reasoning capabilities to handle dynamic and uncertain environments.  Knowledge graphs have emerged as a powerful tool for structuring domain knowledge, enabling symbolic reasoning, and enhancing decision-making in autonomous systems. However, structured knowledge alone may be insufficient for addressing unforeseen challenges and open-ended queries. To bridge this gap, RAG has been proposed as an effective approach that dynamically retrieves and integrates unstructured information into the decision-making pipeline \cite{rag}.

Several works have explored the application of KGs in underwater autonomy. For instance, a semantic world model framework has been proposed to improve interoperability and situation awareness in autonomous underwater systems \cite{5432174}. In addition, ontology-based approaches have been developed for the diagnosis of faults in  AUVs, improving intelligent diagnosis and recovery mechanisms \cite{article}.

Probabilistic graphical models, such as Bayesian Networks and Markov Decision Processes (MDPs), have been employed to handle uncertainty in underwater robotics \cite{li2017towards, hegde2018bayesian}. These models are particularly useful for reasoning under uncertainty and optimizing decision-making in autonomous agents. However, they require significant prior knowledge and may not scale well in complex, data-rich environments.

Deep learning-based approaches, particularly transformer-based models, have recently gained attention for knowledge representation in underwater applications. These models can learn implicit knowledge representations from large datasets, but they often lack interpretability and require extensive labelled data \cite{zhang2022underwater}. Additionally, they struggle with incorporating structured domain knowledge, which limits their effectiveness in highly specialized tasks.

Recent advancements in RAG have demonstrated its capability to enhance AI-generated responses by incorporating relevant external information \cite{li2025enhancingretrievalaugmentedgenerationstudy}. A survey on agentic RAG explores how autonomous AI agents can adapt retrieval strategies to improve response accuracy \cite{singh2025agenticretrievalaugmentedgenerationsurvey}. These techniques provide a promising foundation for integrating RAG into underwater missions, allowing autonomous agents to adaptively retrieve mission-critical knowledge from scientific literature, sensor logs, and operational reports.

By combining KGs and RAG, our approach leverages the strengths of both structured and unstructured knowledge representations. This hybrid system enables robust reasoning while maintaining the flexibility to adapt to novel situations. Compared to purely ontology-based systems, our approach reduces manual curation efforts. In contrast to probabilistic models, it allows for the integration of unstructured, contextual information. Furthermore, compared to deep learning-based approaches, our method retains interpretability while improving adaptability through retrieval mechanisms.

Various control architectures could benefit from the knowledge representations to execute the actions needed to achieve the mission goal. A Finite State Machine (FSM) \cite{kleene1951representationof} is a computational model representing systems through a finite collection of states and definitive transitions triggered by specific inputs or events. It enables clear, predictable, and verifiable control mechanisms. A subsumption architecture \cite{brooks1986robust} is a layered control framework that decomposes robotic behavior into a hierarchy of reactive modules, where lower layers provide essential sensor-driven responses and higher layers can override these behaviors based on contextual priorities. A decision tree \cite{de2013decision} is an algorithmic model that breaks down complex decision-making processes into a structured hierarchy of binary or multi-way splits based on feature thresholds. Although these techniques offer structured solutions for handling decision-making, they may not fully furnish the reactivity required to cope with the unforeseen changes of extremely dynamic settings.

Behaviour Trees (BTs) \cite{dromey2006formalizing, Colledanchise_2018} offer a structured framework for modelling the behaviour of autonomous agents, enabling flexible, scalable, and responsive decision-making processes. They are composed of nodes that represent various elements, such as actions, conditions, and control flows, all organized within a hierarchical tree structure.
Applying dynamic BT \cite{Colledanchise_2019} has been recognized as an important enhancement for handling unpredictable events that were not initially considered. Utilizing the capabilities of Reinforcement Learning (RL), a self-adaptive model that incorporates BTs and RL has been created to enhance decision-making under changing conditions \cite{hu_self_adaptive_2021}. Additionally, a refined Reactive Behaviour Tree technique has been formulated to effectively respond to changes in environments with limited visibility, using cause-effect reasoning to improve system flexibility by either extending BTs with new sub-trees or altering existing ones when condition nodes fail \cite{li2022towards}. Despite these advancements, these methods bring predictability challenges. New decision branches enhance adaptability but hinder users' ability to foresee system actions, making decision-making more complex.

\begin{figure*}[t]
    \centering
    \includegraphics[width=0.7\linewidth]{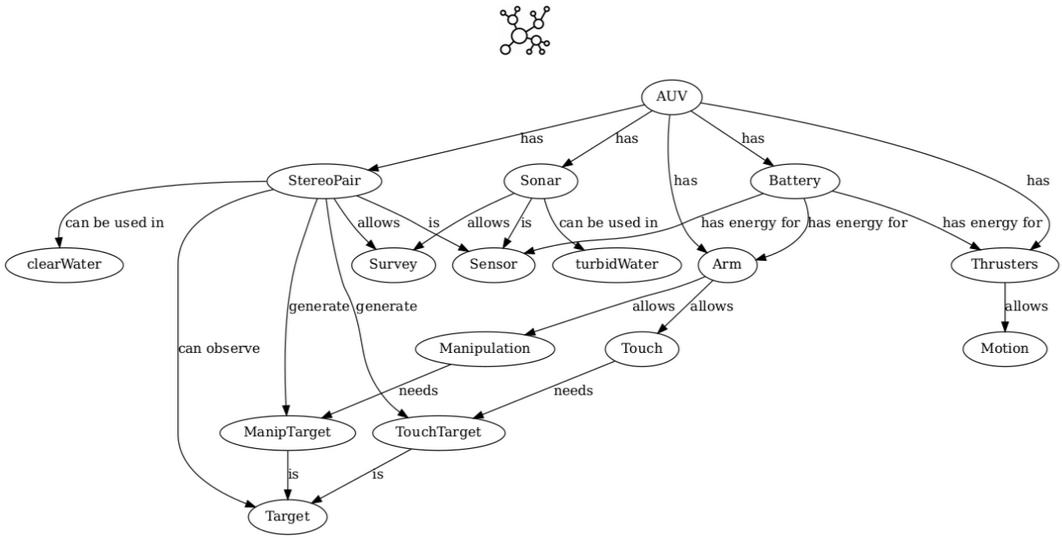}
    \caption{Knowledge Graph: Robot Capabilities, Sensors, and Available Actions.}
    \label{fig:kr}
\end{figure*}

\begin{figure}[htb]
    \centering
    \includegraphics[width=0.8\linewidth]{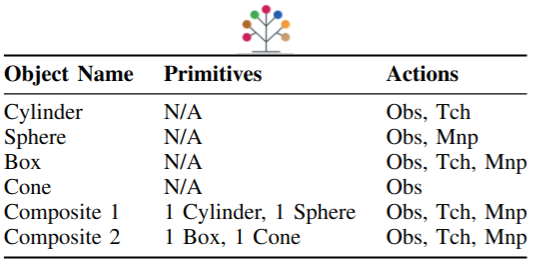}
    \caption{Taxonomy: Composition of Geometric Primitives and
Their Associated Executable Actions.}
    \label{fig:taxonomy}
\end{figure}

\section{Problem Statement}

The goal of this work is to enable shared autonomy in multi-agent marine robotic systems, supporting both independent operation and effective human-robot collaboration. To this end, we propose an architecture that integrates a RAG system---combining a large language model with a structured taxonomy and knowledge graph---to support reasoning and decision-making in autonomous underwater missions.

Specifically, we demonstrate this architecture in two key scenarios: (i) a single AUV (\emph{brain}) system interacting with a human operator, and (ii) a team of two heterogeneous AUVs, $A_1$ (Alpha) and $B_2$ (Beta), communicating and coordinating tasks. 
We assume that both AUVs have different capabilities, such that they need to coordinate in order to perform tasks. For example, $A_1$ has a stereo camera that allows for accurate sensing, while $B_2$ has a monocular camera, but is equipped with a manipulator that allows it to perform intervention tasks. 
We aim to validate the concept of autonomous exploration and detection, where $A_1$ identifies and classifies objects $O = \{o_1, o_2, \dots, o_n\}$ based on sensor data modeled by $f: E \times S \to O$, with $E$ representing the environment and $S$ the sensor readings.
If an object is unknown, the AUV $A_1$ should either: i) enable the human-in-the-loop, or ii) utilize the other AUV $B_2$ to get further information about the objects in the scene. In the human-in-the-loop case, the human classification is enabled via a function $g: O \times H \to T$, where $H$ denotes human input and $T = \{t_1, t_2, \dots, t_m\}$ is a predefined taxonomy. The architecture also supports information exchange between $A_1$ and $B_2$---including object locations $L(o_i)$ and classifications $c(o_i)$---through Visual Light Communication (VLC), modeled by $v: A_1 \times B_2 \times O \to \mathbb{R}$, ensuring real-time coordination.

Finally, the RAG system, informed by the knowledge graph, determines the most appropriate action $a_i \in \{\text{observe}, \text{touch}, \text{collect}\}$ for each object based on its context. This decision-making process is governed by a behavior tree, modeled as $BT: O \times G \to A$, where $G$ represents the knowledge graph and $A$ the action space.

\section{Methodology}  
The increasing reliance on AUVs for complex marine operations highlights the pressing need for effective knowledge representation and decision-making in challenging underwater environments.  
AUVs rely on advanced decision-making capabilities to interpret sensory data, navigate unpredictable terrains, and autonomously respond to changing conditions. However, in particularly harsh environments, an AUV may require guidance from an external operator to determine the best course of action. These autonomous and collaborative abilities form the foundation for achieving diverse goals, such as environmental monitoring, infrastructure inspection, search and rescue, and resource mapping.  
Our methodology enables multi-agent and shared autonomy by integrating knowledge graphs with behavior trees, allowing AUVs to operate independently while still providing actionable insights to human operators when needed. Moreover, we incorporate perception techniques, including point cloud segmentation and compression, to efficiently process and transmit 3D environmental data during missions, ensuring reliable operation even in bandwidth-constrained conditions. To facilitate seamless interaction between the AUV’s onboard systems and the LLM, we deploy a set of ROS1/ROS2 nodes that interface directly with the knowledge graph, structured taxonomy, and various ROS topics. These nodes collect and compress relevant information—including sensor states, battery levels, mission parameters, and environmental observations—into structured JSON files. These files serve as the primary input to the LLM, enabling it to reason over the AUV’s current context in real time. This architecture acts as a communication bridge, ensuring that the LLM operates with coherent, up-to-date situational awareness in order to generate the correct behaviors.

\subsection{Knowledge representation and reasoning}  
We enhance the knowledge representation capabilities of underwater embedded service agents~\cite{patron2009situation,patron2011embedded} by using a RAG system integrating an LLM with a knowledge graph and a structured taxonomy. The knowledge graph provides the AUV with a detailed understanding of its environment and operational capabilities as depicted in Figure \ref{fig:kr}. By incorporating data about its onboard sensors such as imaging devices, environmental sensors, and manipulators the AUV can evaluate its ability to perform specific actions in response to different environmental conditions. For instance, the graph enables the AUV to recognize which sensors are available, determine how to gather information effectively, and decide how to interact with objects in their surroundings.  
Complementing the knowledge graph, which encodes the robot’s internal components and capabilities, a structured taxonomy characterizes object categories in the environment by their constituent geometric primitives and the executable actions they afford. Each object is defined using simple geometric primitives or their combinations, offering clarity on how the AUV should identify and interact with them. The executable actions include: Obs (Observe), Tch (Touch), and Mnp (Manipulate), as shown in the Figure~\ref{fig:taxonomy}.
This structured approach allows the AUV to efficiently interpret its surroundings and execute appropriate tasks. The RAG system enhances decision-making by retrieving relevant information from the knowledge graph and taxonomy, enabling the AUV to generate context-aware actions and responses. Specifically, we used OpenAI's ChatCompletion API using the GPT-3.5-turbo model~\cite{OpenAI2024}. The model processes input based on the current knowledge graph, taxonomy, battery status, and the topic that contains transformations of known and detected objects. Since the knowledge graph and taxonomy can be updated by a human operator, the system refreshes its data every second, ensuring that the AUV operates with the most up-to-date information.

\subsection{Mission Planning and Execution: Behavior Tree}
\begin{figure}[t]
  \centering
  \includegraphics[width=0.8\linewidth]{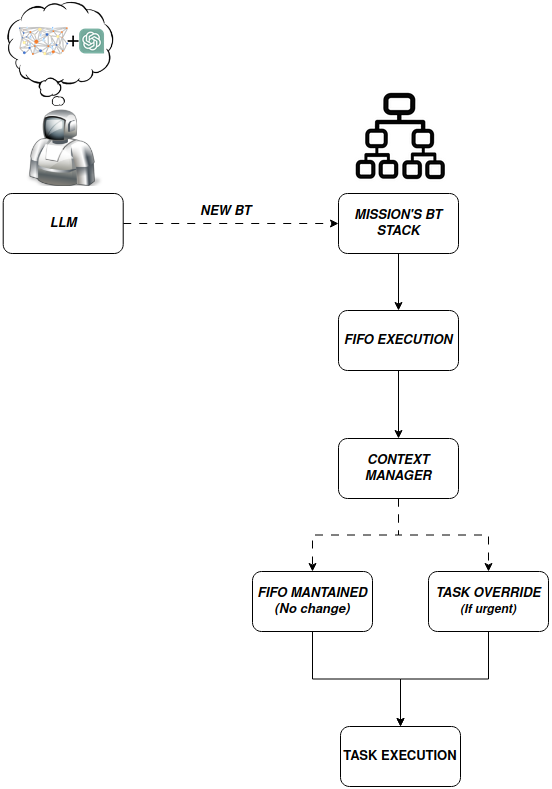}
 \caption{Behavior Tree (BT) management strategy for autonomous decision-making and mission execution in AUVs, with dynamic task prioritization based on real-time conditions.}
  \label{fig:bt_strategy}
\end{figure}

Our methodology integrates BT management with the agent's knowledge representation to enable autonomous decision-making and mission planning. The agent receives a mission file as input, which defines the mission objectives using a ``subject-action'' or ``subject-action-target'' structure. Initially, the robot queries the knowledge graph to determine whether it can accomplish the mission as defined or if an alternative approach is necessary (e.g., utilizing different available sensors).
Once this capability assessment is complete, the RAG system validates the mission using the robot's information from the knowledge graph.
At this stage, the agent instantiates the BTs, which govern the mission's execution.
The BT stack, managed by the LLM within the RAG system, governs the task execution sequence for AUVs. %
 The BT execution is dynamically adapted using a context-aware behaviour-switching strategy as depicted in Figure \ref{fig:bt_strategy}. %
The system continuously evaluates real-time conditions to determine the priority of the task. A context manager monitors key variables, enabling the LLM to decide whether a new BT should override the current sequence based on urgency and relevance. This approach ensures efficient task execution, balancing structured workflows with the flexibility needed for optimal decision-making. 

\section{Case Study}\label{sec:sim}
In this section, we introduce the use case employed to generate our results. We used the Stonefish simulator \cite{cieslak2019stonefish, grimaldi2025stonefishsupportingmachinelearning} to model a multi-agent AUV mission operating under shared autonomy in a submerged environment. In such complex settings, AUVs benefit greatly from human oversight, especially when perception and decision-making are affected by uncertainty. By incorporating a Human-in-the-Loop (HITL) approach, operators are able to intervene in critical processes such as perception, communication, and knowledge enrichment, thereby enhancing mission reliability and adaptability. Both the human operator and the robot can dynamically update the knowledge graph and the taxonomy.

\begin{figure*}[t]
  \centering
  \includegraphics[width=0.95\linewidth]{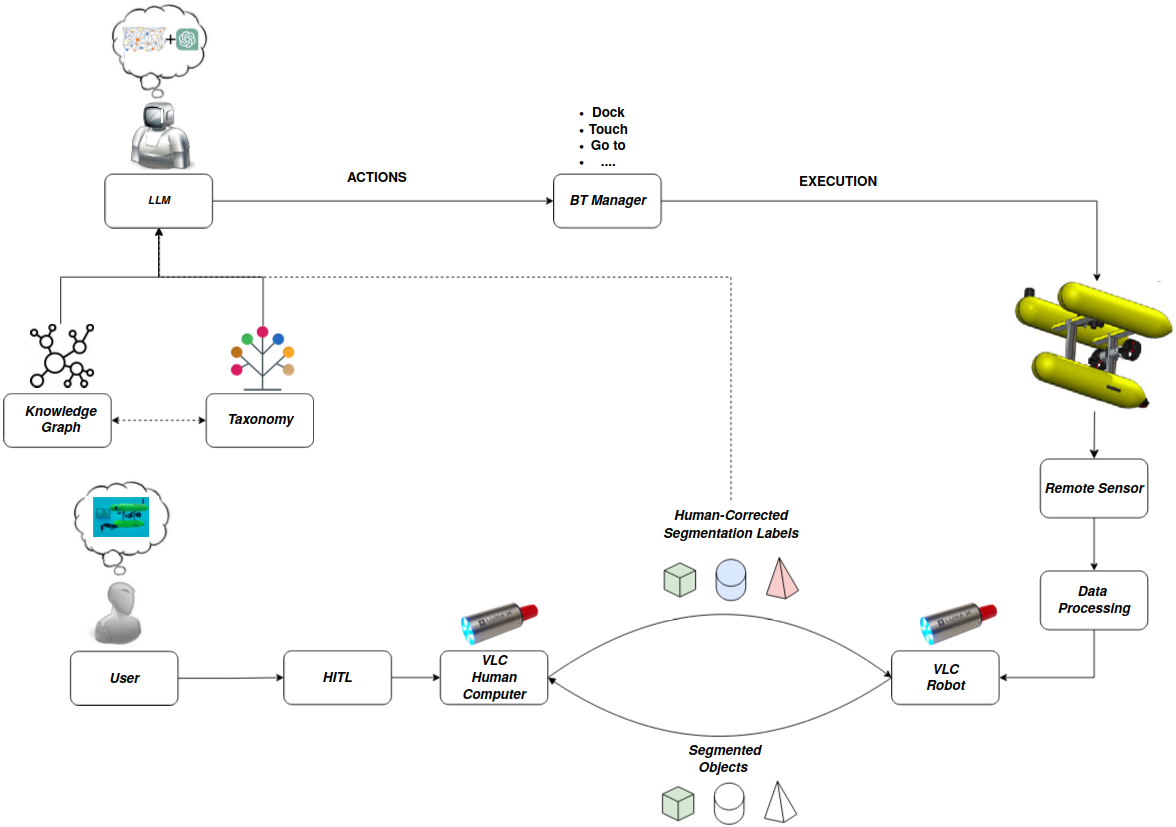}
  \caption{Full and shared autonomy use-case schema. }
  \label{fig:use_case_system_architecture}
\end{figure*}

Figure \ref{fig:use_case_system_architecture} illustrates the autonomy framework, emphasizing the interaction patterns and decision-making flow that characterize shared autonomy. The scenario features two AUVs, each with distinct capabilities and roles:
\begin{itemize}
\item \emph{Alpha} is equipped with RGB and depth cameras and serves primarily as the survey unit.
\item \emph{Beta} is equipped with RGB and depth cameras as well as an underwater manipulator, making it primarily responsible for intervention tasks.
\end{itemize}

\noindent Both robots are also equipped with a VLC sensor that allows them to exchange information with each other. Additionally, the robots also have access to a docking station through which they can use the VLC sensor to communicate with the human operator.  

In this case study, the AUVs are deployed to explore a marine environment in search of specific objects located on the seafloor. Their mission involves surveying the area, detecting and identifying objects within the scene, and, when instructed by the user, interacting with them—either by touching or retrieving. For demonstration purposes, we use simple, easily recognizable objects to facilitate human interpretation. The objects of interest are categorized into basic 3D geometric primitives: (i) sphere, (ii) cylinder, (iii) cube, (iv) cone, and (v) torus. This methodology can be extended to support seafloor mapping tasks, as the one performed in \cite{grimaldi2025realtimeseafloorsegmentationmapping} or any long-term mapping \cite{grimaldi2024fragg}.

\begin{figure}[t]
\centering
\begin{subfigure}[b]{0.20\textwidth}
\centering
\includegraphics [width=\linewidth]{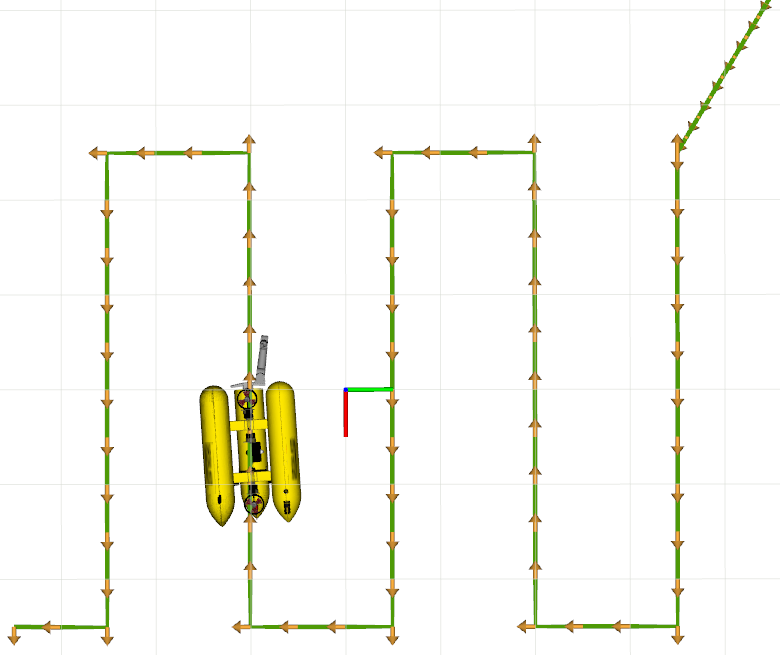}
\end{subfigure}
\hfill
\begin{subfigure}[b]{0.25\textwidth}
\centering
\includegraphics[trim={50 0 0 0},clip,width=0.9\linewidth]{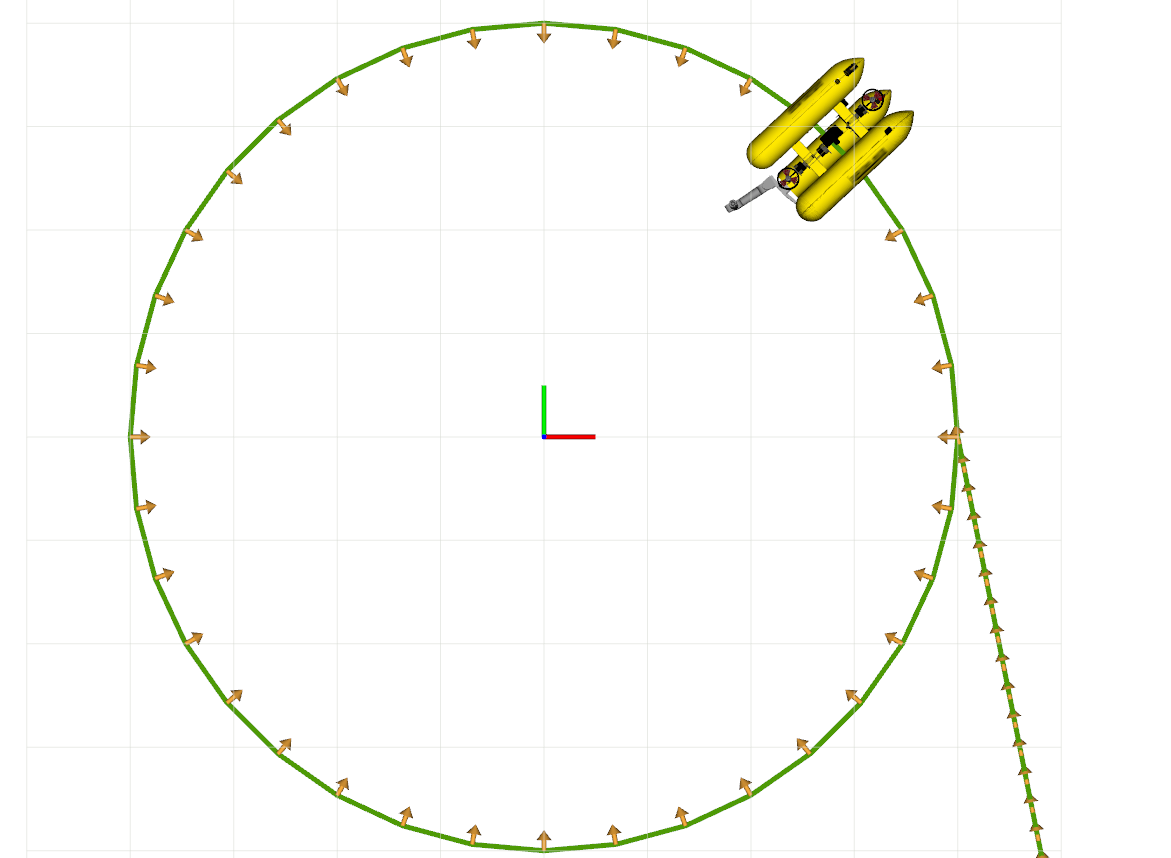}
\end{subfigure}
\caption{Discovery and inspection trajectories generated by the RAG system.}
\label{fig:trajectories}

\end{figure}

In this multi-robot setup, Alpha acts as the explorer, responsible for surveying the environment, detecting objects, and classifying them based on its own sensors. During this process, it surveys the area using simple motion patterns generated by the LLM, shown in Figure \ref{fig:trajectories}, while actively searching for objects. It is important to note that we do not pre-specify any search patterns. Alpha is free to choose and change its search pattern or adapt its behavior if an object is found. 

While this seems like a simple mission, it has a high level of complexity. Both AUVs need to be able to perform their individual tasks while simultaneously managing complex high-level decision-making, which includes when and what information to communicate with the other agents in the team. As such, the successful execution of the mission represents a milestone in marine autonomy. While the presented results are only in simulation, we argue that Stonefish is currently one of the most advanced simulators for marine robotics \cite{aldhaheri2025underwaterroboticsimulatorsreview}.

\begin{figure}[t]
  \centering
  \includegraphics[width=0.95\linewidth]{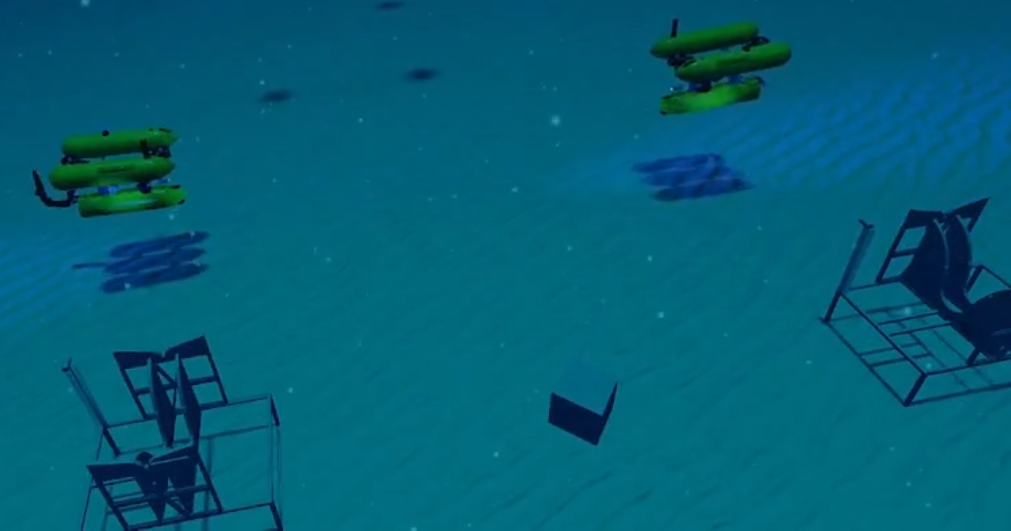}
 \caption{The two AUV using visual-light communication to share knowledge.}
  \label{fig:two_auv}
\end{figure}

\section{Results}

\begin{figure*}
        \centering %
        \begin{subfigure}{0.33\textwidth}
          \includegraphics[width=\linewidth,height=4cm]{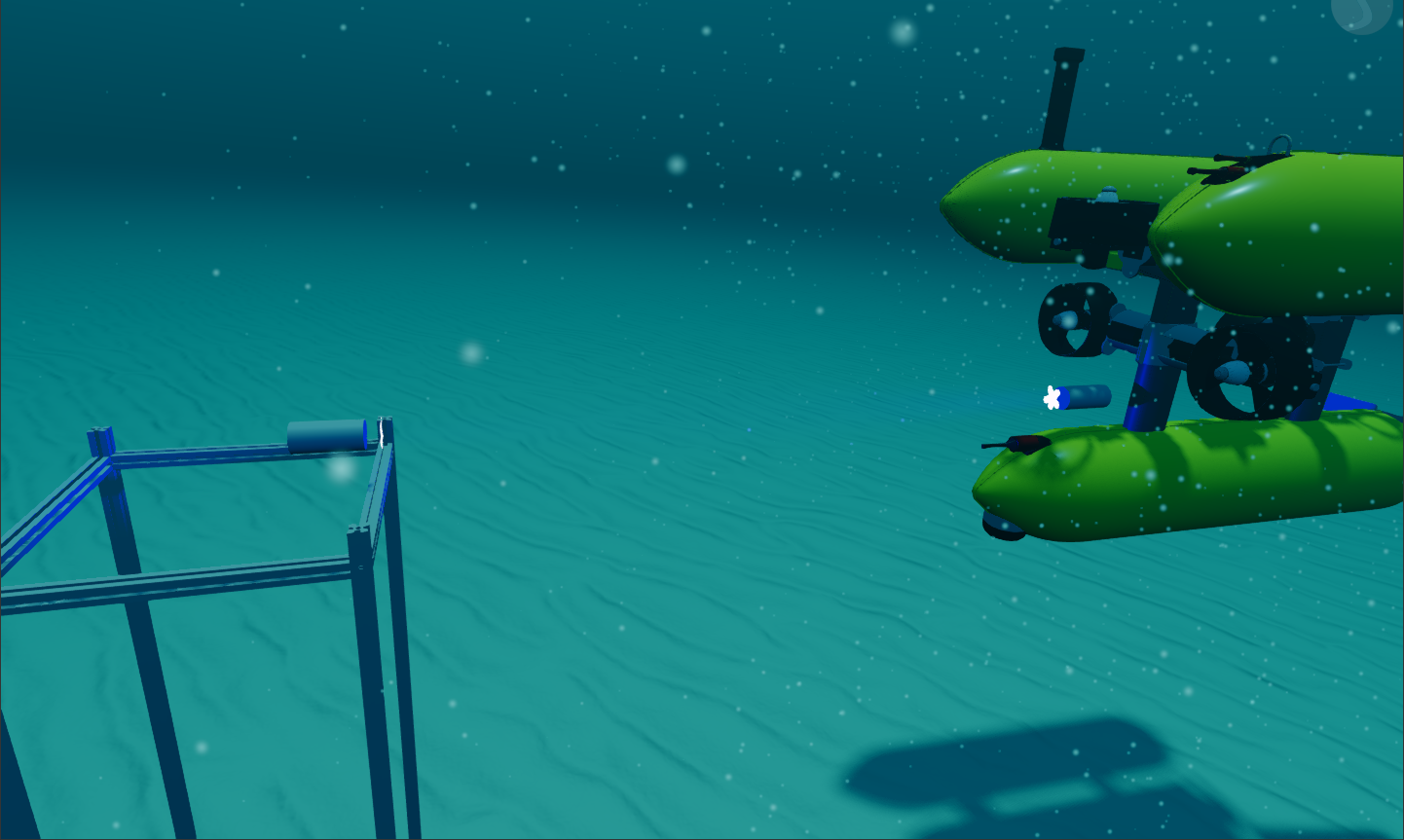}
          \caption{VLC}
          \label{fig:vlc_clean}
        \end{subfigure}\hfil %
        \begin{subfigure}{0.33\textwidth}
          \includegraphics[width=\linewidth,height=4cm]{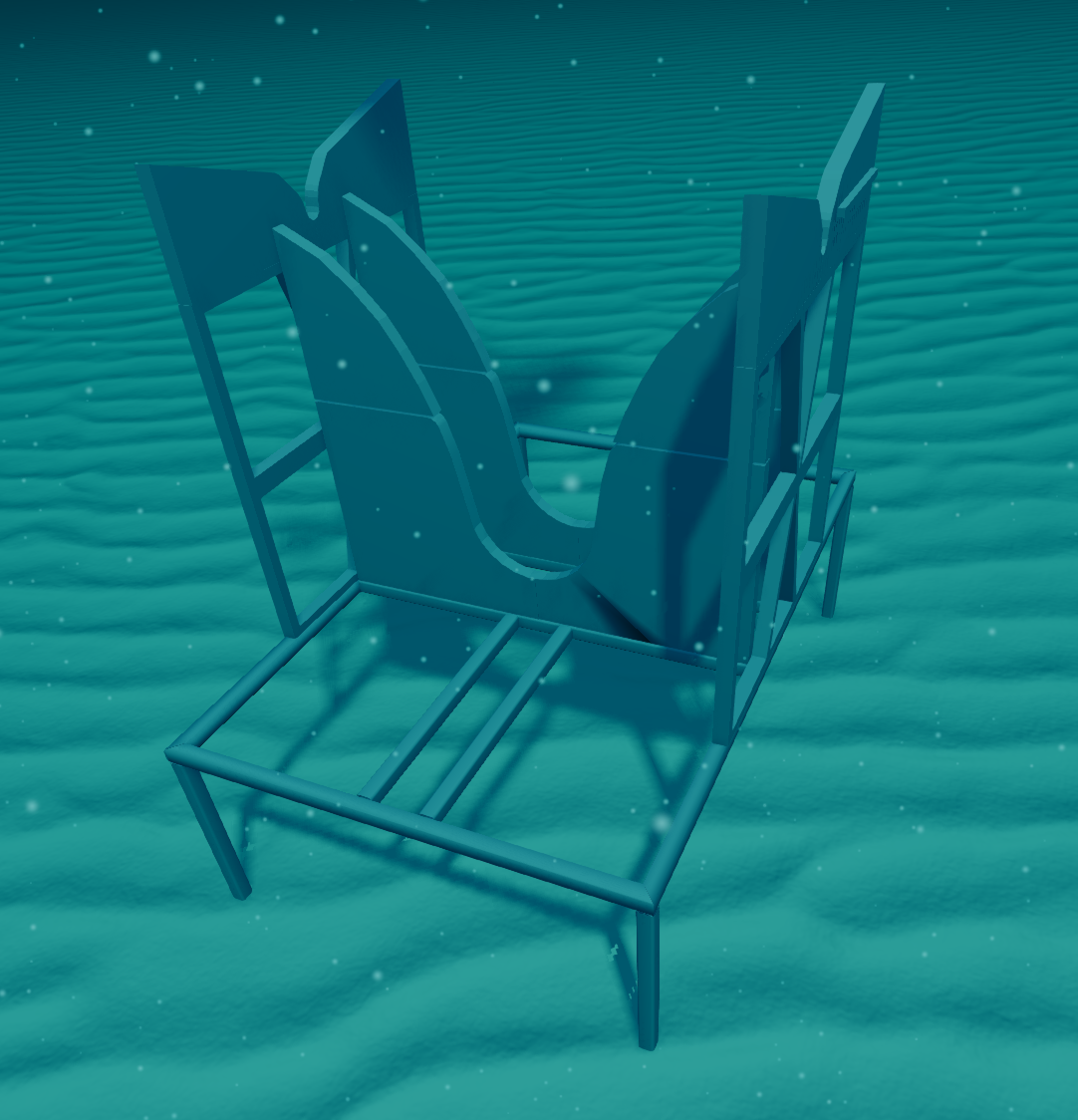}
          \caption{Docking station}
          \label{fig:ds}
        \end{subfigure}\hfil %
        \begin{subfigure}{0.33\textwidth}
          \includegraphics[width=\linewidth,height=4cm]{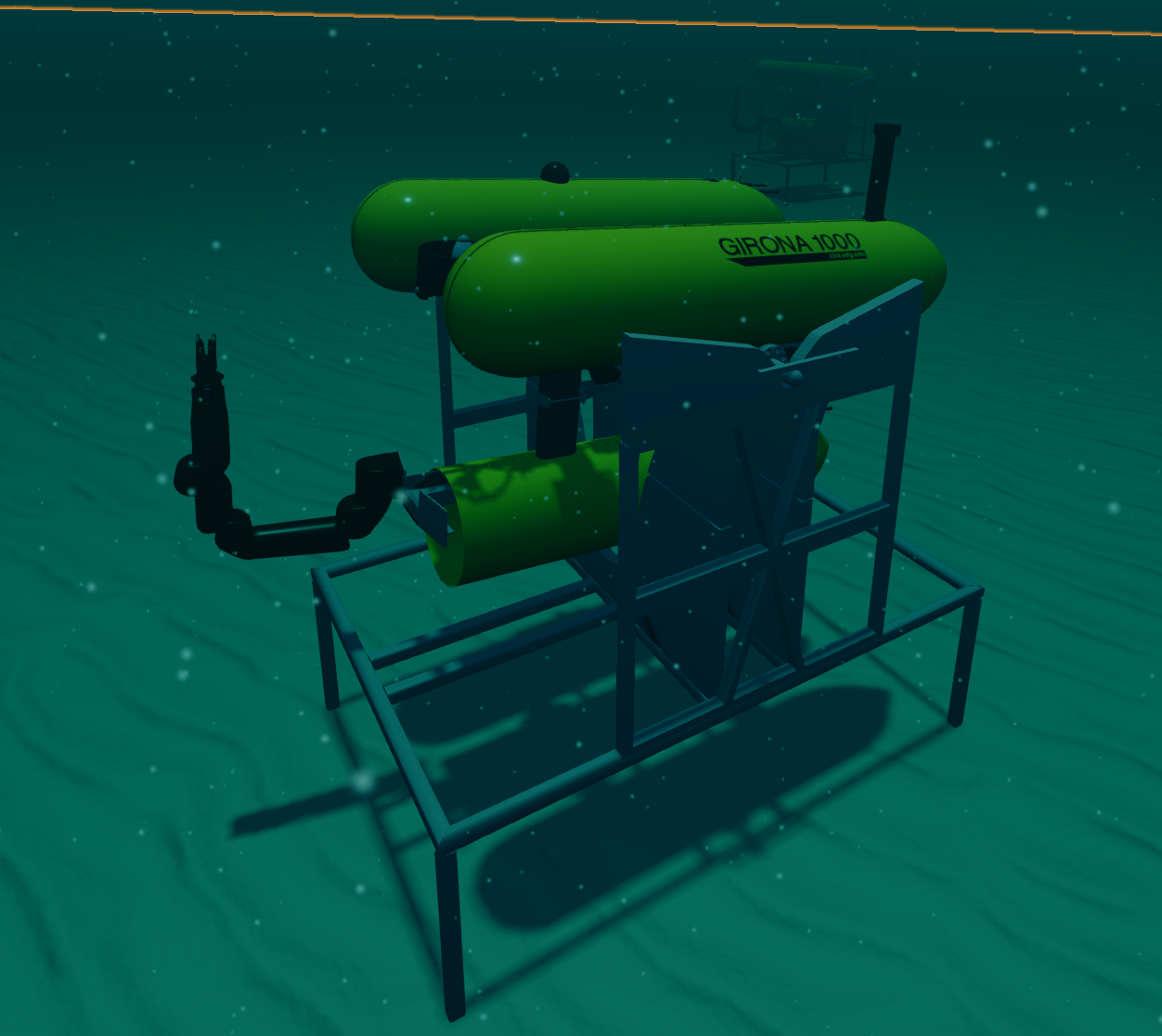}
          \caption{AUV in Docking station}
          \label{fig:auv_in_ds}
        \end{subfigure}\hfil %
         \vspace{0.5cm}
          \includegraphics[width=0.45\linewidth]{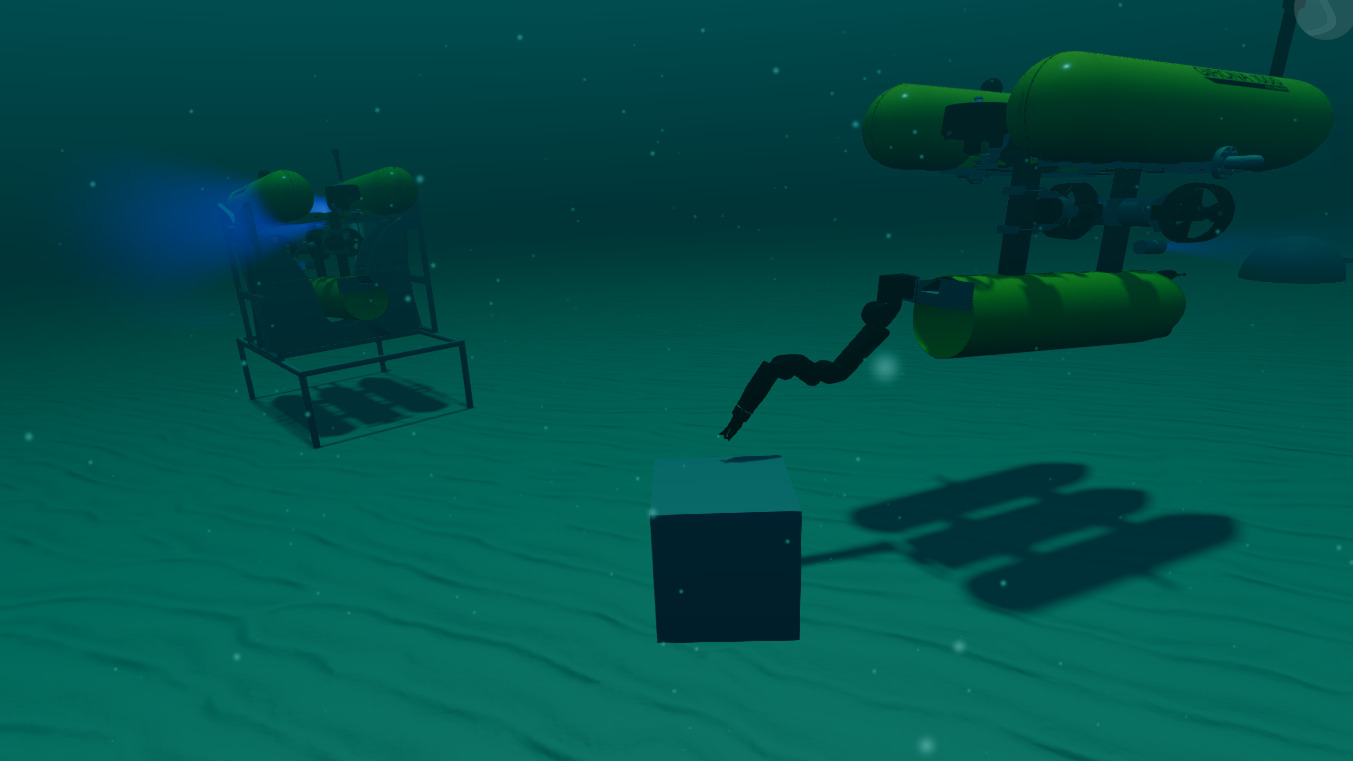}
    \caption{Operational environments for the AUV: (a) VLC mounted on the docking station for communication, (b) docking station setup, and (c) the AUV docked for recharging or maintenance. Bottom: AUV \textit{Beta} interacting with an object, while \textit{Alpha} is recharging at the docking station.}
    \label{fig:ds_fig}
\end{figure*}

\begin{table*}[h]
\centering
\normalsize
\begin{tabular}{|l|c|c|}
\hline
\textbf{Method} & \textbf{Mission Validation Success Rate} & \textbf{BT Completeness} \\
\hline
LLM \& Runtime State + Knowledge Graph + Taxonomy & 100\% & 100\% \\
LLM \& Runtime State  + Knowledge Graph & 85\% & 76\% \\
LLM \& Runtime State & 21\% & 15\% \\
\hline
\end{tabular}
\caption{Comparison of RAG system configurations in terms of mission validation success and Mission Completeness (BT creation per task). Results are based on 20 missions per configuration.}
\label{tab:rag_performance}
\end{table*}

The evaluation of the proposed RAG-based system focused on its effectiveness in mission validation and BT planning within a simulated multi-AUV operational scenario. A demonstration video showcasing the use case has been made available online\footnote{The video is available here: \url{https://michele1996.github.io/rag_full_shared_autonomy.github.io/}}
To assess the contribution of each system component, we conducted an ablation study using the following configurations, each of them having access to the runtime state information (robot state and perceived environment):

\begin{itemize}
    \item \textbf{LLM \& Runtime State + Knowledge Graph + Taxonomy:} Full integration of structured domain knowledge and semantic task constraints.
    \item \textbf{LLM \& Runtime State + Knowledge Graph:} Language model reasoning grounded in factual knowledge without predefined semantic structuring.
    \item \textbf{LLM \& Runtime State:} Pure language-based generation with no semantic knowledge integration.
\end{itemize}

Each configuration was evaluated on 20 distinct missions to measure its mission validation success rate and behavior tree (BT) completeness (i.e., whether a valid BT was created for each task). 
No external baselines were used, as this work introduces a novel architecture that is able to perform tasks not achievable by other works in the state of the art. As such, there are no directly comparable prior systems. Instead, the ablation study serves to isolate and quantify the impact of each knowledge integration component.
The performance metrics for each configuration are summarized in Table~\ref{tab:rag_performance}.

\subsection{Mission execution} 
The scenario starts with the mission for robot Alpha sent by the user via a simulated Wi-Fi antenna.
Prior to execution, the AUV's mission was validated by its respective RAG system, which leveraged the knowledge graph to verify that the robot possessed the necessary capabilities to accomplish its assigned tasks. Upon validation, Alpha's LLM, featuring the Semantic Knowledge and Runtime State, reasoned that the best approach would be to execute a lawnmower-style survey of the area and created the corresponding behavior tree. Followed by a second behavior tree for the reporting phase, where the AUV approached the human operator to relay performance feedback. After completing the survey, Alpha navigated to the docking station to align its VLC system for data exchange.

To stimulate collaborative behavior, we intentionally configured Alpha with a low battery level and introduced partial errors in its object segmentation module. As a result, when an object was incorrectly segmented and Alpha lacked sufficient power to continue, Beta was deployed by the human from its docking station. Its first mission is to communicate with Alpha and receive the information collected and the human feedback. In this case, Alpha shares a task with Beta: to inspect an object that was previously misclassified. Beta’s RAG system validates the initial part of the mission, and both AUVs are then instructed by their respective RAG systems to move to precise positions. This positioning enables Alpha to transfer its data to Beta via VLC, as illustrated in Figure \ref{fig:two_auv}. Once the data transfer was completed, Alpha was instructed to return to the docking station to conserve and restore power. Subsequently, Beta's RAG validated the second part of the mission and instructed it to proceed with the inspection and circumnavigation of the misidentified object, performing a refined segmentation to correct the error.

\subsection{Mission Validation Accuracy}
One of the most consistent and impactful findings was the reliability of the full RAG configuration in mission validation. The system was able to correctly assess whether a task was feasible for a given AUV, never assigning actions that exceeded vehicle capabilities. This outcome can be directly attributed to the incorporation of the knowledge graph, which provided detailed specifications of vehicle parameters such as sensor ranges, manipulation capabilities, battery endurance, and depth limits. The predefined taxonomy further enhanced this validation process by enforcing semantic alignment between tasks and AUV roles, preventing mismatches like assigning manipulation tasks to non-manipulator vehicles.

By contrast, the LLM-only configuration exhibited frequent failures in validation. Lacking awareness of operational constraints, the LLM often overestimated AUV capabilities or hallucinated nonexistent ones. For instance, it occasionally assigned ``object retrieval'' missions to vehicles without manipulators or planned high-depth surveys for shallow-water units. These errors underline the limitations of relying solely on language priors without factual or semantic grounding.

The LLM+KG configuration partially mitigated these failures by anchoring generation in accurate vehicle data. However, without the taxonomy, the system sometimes misapplied tasks to objects in ways that were logically plausible in general terms but semantically invalid according to the operational context. For example, the system might assign an ``inspect'' task to a cylindrical object that, per the predefined taxonomy, is not an object type meant to be inspected, whereas a cube might be considered a valid inspection target. These object-task relationships are defined explicitly in the taxonomy, and without it, the LLM lacked a clear mapping between object types and valid actions. This resulted in plausible-sounding but operationally inappropriate plans. 

\subsection{Planning via Behavior Trees}
In addition to validating tasks, the system was evaluated on its ability to produce complete and coherent Behavior Trees (BTs) for each mission. The fully integrated RAG system demonstrated high performance in this area, generating well-structured BTs that included all required subtasks and actions, correctly sequenced. For example, mission plans involving underwater inspection included the full sequence of ``survey'', ``navigate to target'', ``inspect target'', ``doc'' and ``undock'' with appropriate condition checks embedded at each step.

When the taxonomy was removed, the BTs produced (LLM+KG configuration) were more error-prone. Incomplete or redundant steps became more common, particularly in complex multi-phase tasks such as object interaction or area mapping. The taxonomy served as a blueprint for assembling BTs, providing reusable action templates, and ensuring structural consistency across similar tasks.

The LLM-only setup produced the least reliable BTs. Many were missing critical actions or contained illogical transitions (e.g., attempting to verify a condition after executing the dependent action). In several cases, the BTs violated AUV constraints entirely, planning for unsupported behaviors or incompatible task sequences. These failures further reinforce the need for external grounding in both factual and semantic dimensions.

Overall, the experimental results demonstrated a clear hierarchy of performance across the three configurations:

    The \textbf{full RAG system} achieved high reliability in both mission validation and plan generation, consistently producing feasible, executable tasks and coherent BTs.

    The \textbf{LLM+KG system} performed reasonably well in feasibility checks but struggled with structured planning due to the absence of semantic task templates.

    The \textbf{LLM-only system} exhibited frequent errors in both validation and planning, with high rates of hallucination and logical inconsistencies.

These findings underscore the necessity of structured, domain-specific grounding—through both factual knowledge graphs and semantic taxonomies—for robust decision-making in autonomous mission planning.

\section{Conclusion}
In this paper, we introduce a framework designed to enhance the autonomy of underwater robotic systems in multi-robot scenarios, enabling each AUV to independently reason and make decisions. The framework combines an LLM with RAG, and integrates a knowledge graph with a predefined simple object taxonomy. This architecture also utilizes Behavior Trees managed by the RAG system, allowing the AUVs to do mission planning. The proposed approach enables AUVs to autonomously plan and execute missions while dynamically incorporating real-time updates from a knowledge graph, taxonomy files, and mission-relevant topics such as battery status and detected object transformations. By structuring decision-making through BTs, the system ensures efficient and flexible mission execution, adapting to environmental uncertainties while maintaining a structured operational workflow. A key feature of our system is the structured yet flexible human-in-the-loop mechanism, allowing operators to intervene at strategic decision points. This enhances mission reliability while reducing the need for constant supervision, striking a balance between autonomy and control. Our results show that our proposed architecture enables robots to perform complex missions in the marine domain, while at the same time maintaining accuracy. This was demonstrated in our results, where our system was able to successfully perform the task 100\% of the time.

In future work, we aim to evaluate the proposed approach in real-world underwater deployments, similar to those described in \cite{6942870}. Additionally, we plan to extend the range of the AUV's missions, including complex objects and complex tasks.
Moreover, our system provides a foundation for evaluating the autonomy levels of robotic operations by tracking mission execution and human intervention frequency. This quantitative framework allows for assessing the system's efficiency and adaptability under varying environmental and operational conditions.

\section*{Acknowledgment}

This work was partially supported by the EPSRC project UNderwater IntervenTion for offshore renewable Energies (UNITE), grant number EP/X024806/1; the HORIZON project Automated Inspection Robots for Surface, Aerial and Underwater Substructures (AEROSUB), grant number PID-101189723; and the COOPERAMOS project (PID2020-115332RB-C32), funded by the Spanish Ministerio de Ciencia, Innovación y Universidades.

\bibliographystyle{ieeetr}
\bibliography{references}

\end{document}